\documentclass[letterpaper]{article} % DO NOT CHANGE THIS
\usepackage{aaai2027}  % DO NOT CHANGE THIS
\usepackage[hyphens]{url}  % DO NOT CHANGE THIS
\usepackage{graphicx} % DO NOT CHANGE THIS
\usepackage{natbib}  % DO NOT CHANGE THIS AND DO NOT ADD ANY OPTIONS TO IT
\usepackage{caption} % DO NOT CHANGE THIS AND DO NOT ADD ANY OPTIONS TO IT
\usepackage{algorithm}
\usepackage{algorithmic}
\usepackage{amsmath}
\usepackage{amssymb}
\usepackage{newfloat}
\usepackage{listings}
\DeclareCaptionStyle{ruled}{labelfont=normalfont,labelsep=colon,strut=off} % DO NOT CHANGE THIS
\floatstyle{ruled}
\newfloat{listing}{tb}{lst}{}
\floatname{listing}{Listing}

\usepackage{booktabs}

\usepackage{tabularx}

\title{DreamTrajectory: Trajectory-Guided Action Generation with World Model Alignment for Mobile Manipulation}
\author{
    Zheng Yang\textsuperscript{\rm 1}\equalcontrib,
    Wenjie Zhang\textsuperscript{\rm 1}\equalcontrib,
    Xiangyu Chen\textsuperscript{\rm 1},
    Wenxuan Song\textsuperscript{\rm 1},
    Xianpeng Wang\textsuperscript{\rm 1},
    Yihang Kang\textsuperscript{\rm 1},
    Jiawen Wen\textsuperscript{\rm 1},
    Wen Chen\textsuperscript{\rm 2},
    Lujia Wang\textsuperscript{\rm 1},
    Renjing Xu\textsuperscript{\rm 1},
    Haoang Li\textsuperscript{\rm 1},
    Xiaowen Chu\textsuperscript{\rm 1}
}
\affiliations{
    \textsuperscript{\rm 1}Hong Kong University of Science and Technology (Guangzhou)\\
    \textsuperscript{\rm 2}Ola Dimensions\\
}
\begin{document}

\nocopyright
\maketitle

\begin{abstract}
Mobile manipulation requires a robot to coordinate base and arm motion under continuously changing viewpoints and contact conditions, within an action space far larger than that of fixed-base manipulation. Existing Vision-Language-Action (VLA) policies are limited in two respects. (i)~They map observations directly to whole-body action chunks, searching this large action space without an explicit task-space motion plan, which makes coordinated base--arm prediction imprecise. (ii)~They execute the predicted chunk open-loop, without checking whether the actions can realize the motion the policy intended, so control errors and unmodeled contacts accumulate into a gap between planned and realized motion. We present \textbf{DreamTrajectory}, a trajectory-guided framework for language-conditioned mobile manipulation that introduces one component for each limitation. Addressing~(i), DreamTrajectory jointly predicts an intention-level end-effector trajectory and a whole-body action chunk in a single action expert, so that the trajectory explicitly guides base--arm action generation instead of remaining implicit. Addressing~(ii), a lightweight trajectory world model predicts the trajectory that a candidate action chunk would induce, and a test-time search--predict--score procedure selects the candidate best aligned with the planned trajectory. On MS-HAB, trajectory guidance raises average success from $32.3\%$ to $47.5\%$ and test-time refinement further to $54.8\%$, with the largest gains on contact-rich articulated-object tasks. On three real-world mobile manipulation tasks, the corresponding average success rates are $63.3\%$, $81.7\%$, and $90.0\%$.
\end{abstract}

% 轨迹本身也能指导（causal attn），也要写现有方法在无轨迹上，没有check能力，无法保证精度高
% 先写联合预测轨迹，能用轨迹检测合理性，并且做必要修复，我们是一个两阶段的，先。。。再。。。

\begin{figure*}[t]
    \centering
    \includegraphics[width=\textwidth]{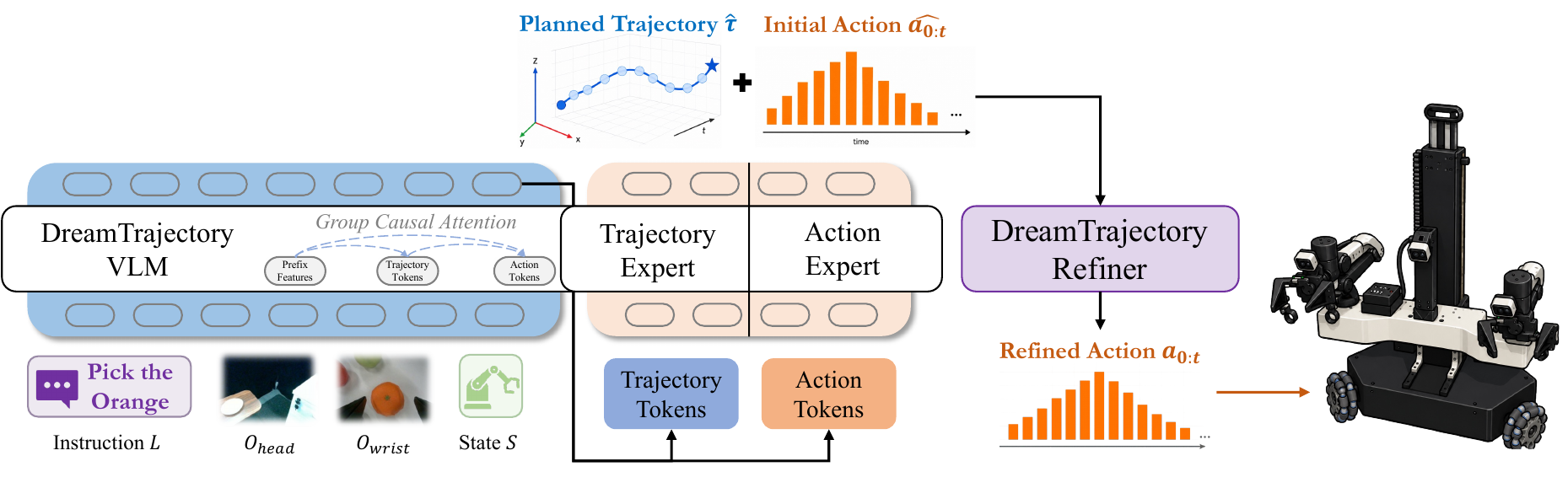}
    % \caption{
    %     Overview of \textbf{DreamTrajectory}. A trajectory-guided VLA jointly predicts an intended
    %     end-effector trajectory and a whole-body action chunk, while a
    %     trajectory world model evaluates and refines candidate actions at
    %     test time.
    % }
    \caption{
        Overview of \textbf{DreamTrajectory}. A trajectory-guided VLA jointly predicts a planned end-effector trajectory and an initial whole-body action chunk. At test time, the DreamTrajectory Refiner refines the initial action chunk before execution based on the planned trajectory.
    }
    \label{fig:dreamtrajectory}
\end{figure*}

\section{Introduction}
\label{sec:introduction}

Mobile manipulation integrates locomotion and dexterous interaction in
unstructured indoor environments, enabling robots to operate well beyond the
workspace of a fixed base.
Unlike fixed-base manipulation, a mobile manipulator must jointly decide how to
coordinate its arm motions and base motions while the camera viewpoint changes
continuously.
These coupled decisions define a substantially larger effective action space
than manipulation alone, and they amplify sensitivity to execution error: a
small deviation early in a whole-body action chunk shifts both the contact
state and the observation from which the next chunk is predicted.
As a result, policies that perform well on tabletop benchmarks frequently
degrade when required to execute contact-rich mobile skills such as opening a
refrigerator while adjusting the base pose or closing a drawer with the
assistance of a mobile base.

Existing methods address only part of these challenges.
% TODO: add citations for navigation-then-manipulation decomposition and for
% separately trained base and arm controllers.
A widely used strategy is to factorize mobile manipulation into navigation
followed by manipulation, or to train separate controllers for the base and
arm.
Such decomposition simplifies optimization and interface design, but it restricts behaviors that require tight temporal coupling between locomotion and manipulation, including reaching while translating, manipulating articulated objects during base repositioning, and recovering from contacts that depend on whole-body compliance.

End-to-end
VLAs~\cite{kim2024openvla,black2024pi_0,bjorck2025gr00t,intelligence2025pi_}
avoid explicit decomposition and can, in principle, represent coordinated
whole-body behaviors, yet they remain limited in two respects.
First, they map observations directly to whole-body action chunks and therefore
search a high-dimensional action space without an explicit task-space motion
plan; the motion that a chunk is meant to realize stays implicit, which makes
coordinated base--arm prediction imprecise.
Second, they execute the predicted chunk open-loop: no mechanism verifies
whether the commanded actions will actually produce the intended motion, so
control errors, collisions, and out-of-distribution contacts are revealed only
after the behavior has already failed.
These limitations motivate \textbf{DreamTrajectory} we propose, a trajectory-guided VLA framework for
language-conditioned mobile manipulation.
Our central hypothesis is that a compact end-effector trajectory provides a
physically meaningful intermediate representation for whole-body control: it
specifies \emph{where} the gripper should move before \emph{how} base and arm
commands should realize that motion.

To address the first limitation, DreamTrajectory generates the planned
trajectory and the whole-body action chunk jointly inside a single action
expert.
Given egocentric head-camera and wrist-camera images, proprioceptive state, and
a language instruction, the two streams are denoised synchronously under a
group-causal attention mask, so that the evolving trajectory stream conditions
action denoising while action tokens cannot influence the trajectory.
The trajectory therefore serves as a task-space reference for guiding action generation in the large whole-body action space.

To address the second limitation, DreamTrajectory introduces a lightweight
\emph{action-conditioned trajectory world model}, which
predicts, from the current observation, the end-effector trajectory that a
candidate action chunk would induce.
This mapping is learned from interaction data rather than derived analytically,
because the deviation between commanded and realized motion originates from
contacts with external objects, self-collisions of the arm, and tracking error
of the low-level controller, none of which an analytic dynamics model captures
reliably.
At test time, DreamTrajectory performs \emph{search--predict--score}
refinement: it samples action candidates around the initial proposal, predicts
the trajectory induced by each candidate, and selects the one that best agrees
with the planned trajectory while remaining smooth.
Because candidates are evaluated in a compact trajectory space rather than in
pixel space, the world model uses only around $49$M parameters and scores $N$
candidates in parallel, and can be attached to the trajectory-guided policy
as a plug-in stage that requires no retraining.
This provides a test-time consistency mechanism, since candidates whose
predicted execution deviates from the plan can be down-ranked before execution. Table~\ref{tab:qualitative_comparison} positions DT with respect to two related
directions: Trajectory-guided policies provide intermediate motion references,
whereas action-conditioned world models predict the consequences of candidate
actions. DT connects them by jointly generating a planned trajectory and
whole-body actions, predicting each candidate's induced trajectory, and
explicitly aligning the two before execution.

We evaluate DreamTrajectory on six \texttt{set\_table} subtasks of the MS-HAB
benchmark~\cite{shukla2025maniskill} with a Fetch mobile manipulator, and on
three real-world pick-and-place and drawer-manipulation tasks with a physical ARX LIFT
platform.
Trajectory guidance raises mean subtask success from $32.3\%$ to $47.5\%$ over
an action-only $\pi_{0.5}$ baseline, and test-time refinement brings it to
$54.8\%$.
On the real platform, trajectory guidance raises average success from $63.3\%$
to $81.7\%$, and refinement further raises it to $90.0\%$.

Our contributions are as follows:
\begin{itemize}
    \item We formulate mobile manipulation action generation as the joint
    generation of a planned end-effector trajectory and a whole-body action
    chunk within a single action expert, where a group-causal attention mask
    turns the trajectory into an explicit task-space guide for coordinated
    base--arm control.
    \item We introduce a lightweight action-conditioned trajectory world model together
    with a search--predict--score procedure that selects, before execution, the
    candidate action chunk whose induced trajectory best matches the plan.
    Operating in trajectory space rather than pixel space, it is a lightweight
    test-time stage that requires no retraining of the policy.
    \item We demonstrate consistent improvements on six MS-HAB
    \texttt{set\_table} subtasks, raising mean success from $32.3\%$ to
    $54.8\%$, and on three real-world tasks, raising average success from
    $63.3\%$ to $90.0\%$ on a physical mobile manipulator.
\end{itemize}

\begin{figure*}[!t]
    \centering
    \includegraphics[width=\textwidth]{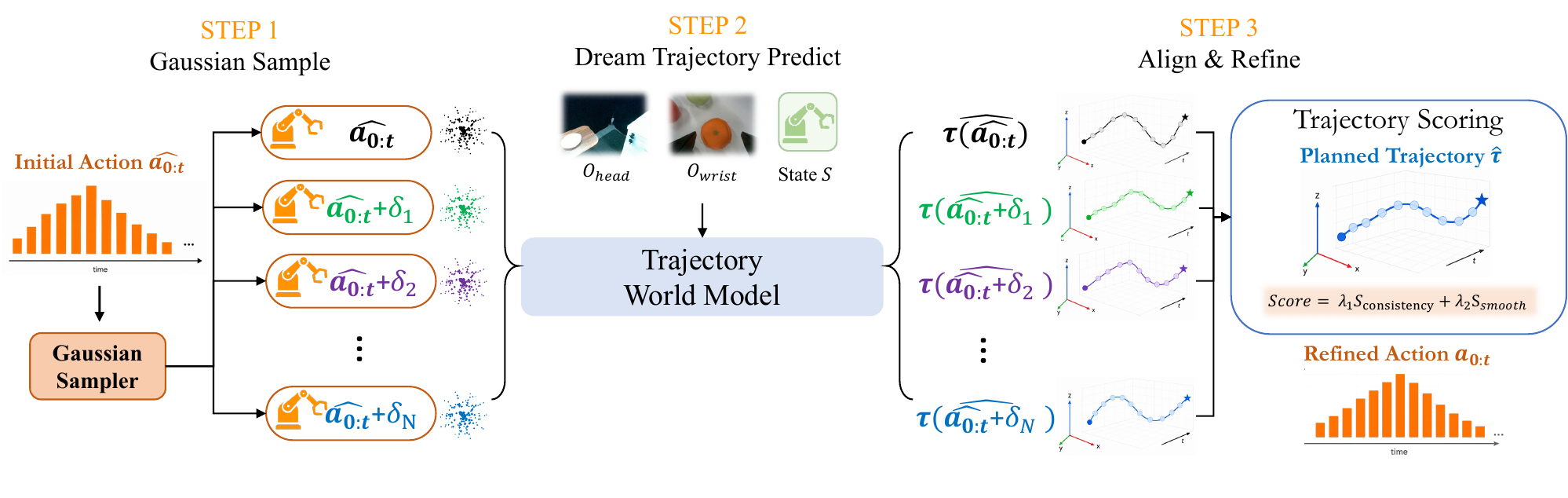}
    % \caption{
    %     Test-time action refinement in \textbf{DreamTrajectory}. The refinement module takes the initial action and the planned trajectory as input. Action candidates near the initial action are sampled and
    %     rolled out by the trajectory world model, and are then ranked according to their
    %     consistency with the planned trajectory.
    % }
    \caption{
Test-time action refinement in \textbf{DreamTrajectory}.
Candidate actions are sampled around the initial action, and the trajectory
world model predicts their induced trajectories. The candidates are ranked according to planned–induced trajectory consistency and action smoothness, and the highest-scoring candidate is executed.
}
    \label{fig:refine}
\end{figure*}

\begin{table*}[t]
\centering
\small
\setlength{\tabcolsep}{4pt}
\renewcommand{\arraystretch}{1.08}
\caption{Methodological comparison of trajectory-guided policies, action-conditioned world models, and DreamTrajectory. The comparison concerns methodology rather than quantitative
performance.}
\label{tab:qualitative_comparison}
\begin{tabularx}{\textwidth}{
@{}
p{2.5cm}
p{3.0cm}
>{\raggedright\arraybackslash}X
>{\centering\arraybackslash}p{1.2cm}
>{\centering\arraybackslash}p{2.0cm}
@{}
}
\toprule
Method
& Future Representation
& Role in Action Generation or Control
& Mobile
& Plan--Execution Alignment \\
\midrule
\multicolumn{5}{@{}l}{\textit{Trajectory-guided policies}}\\
DTP~\cite{fan2025diffusion}
& 2D pixel trajectory
& Separate guidance for policy learning
& $\times$ & $\times$ \\
MoManipVLA~\cite{wu2025momanipvla}
& 3D pose trajectory
& External input for base--arm optimization
& \checkmark & $\times$ \\
\midrule
\multicolumn{5}{@{}l}{\textit{Action-conditioned world models}}\\
DINO-WM~\cite{zhou2024dino}
& Latent
& Rollout for planning
& $\times$ & $\times$ \\
V-JEPA-AC~\cite{assran2025v,mur2026v}
& Latent
& Prediction for planning
& $\times$ & $\times$ \\
\midrule
\multicolumn{5}{@{}l}{\textit{Integrated trajectory guidance and prediction}}\\
DreamTrajectory
& 3D pose trajectory
& Joint generation and test-time refinement
& \checkmark & \checkmark \\
\bottomrule
\end{tabularx}
\end{table*}

\section{Related Work}
\label{sec:related}

\subsection{Generalist Vision-Language-Action Policies}
\label{sec:related:vla}

RT-1~\cite{brohan2022rt}, RT-2~\cite{zitkovich2023rt}, and
PaLM-E~\cite{driess2023palm} established scalable vision-language robot
learning, while Open X-Embodiment and RT-X~\cite{o2024open},
Octo~\cite{team2024octo}, and RoboFlamingo~\cite{li2024vision} explored
cross-task and cross-embodiment transfer.
More recent VLAs, including OpenVLA~\cite{kim2024openvla},
RDT-1B~\cite{liu2025rdt}, the $\pi$ family~\cite{black2024pi_0,
intelligence2025pi_,torne2026mem,intelligence2026pi},
GR00T N1~\cite{bjorck2025gr00t}, CogACT~\cite{li2024cogact},
ReconVLA~\cite{song2026reconvla}, and Unified Diffusion VLA~\cite{chen2026unified},
combine pretrained vision-language representations with specialized action
decoders.
OpenVLA-OFT~\cite{kim2025fine}, SmolVLA~\cite{shukor2025smolvla},
TinyVLA~\cite{wen2025tinyvla}, SpatialVLA~\cite{qu2025spatialvla},
Spatial Forcing~\cite{li2026spatial}, and PD-VLA~\cite{song2025pd}
further improve adaptation efficiency and spatial grounding.
These methods primarily predict action sequences directly, whereas
DreamTrajectory (DT) introduces an explicit end-effector trajectory to guide
whole-body action generation.
\subsection{Mobile and Trajectory-Aware VLA Policies}
\label{sec:related:trajectory_mobile}

Trajectory-aware policies introduce intermediate motion representations beyond
direct action prediction.
RT-Trajectory~\cite{gu2023rt} uses trajectory sketches,
3D Diffuser Actor~\cite{ke20243d} predicts 3D pose trajectories, and
TraceVLA~\cite{zheng2025tracevla}, Diffusion Trajectory-guided
Policy~\cite{fan2025diffusion}, and CoT-VLA~\cite{zhao2025cot} employ visual
traces, image-plane paths, or future images.
GR-2~\cite{cheang2024gr}, DreamVLA~\cite{zhang2026dreamvla}, and
WorldVLA~\cite{cen2025worldvla} more tightly couple future prediction with
action generation.

Mobile manipulation additionally requires coordinated base--arm control under
changing viewpoints and whole-body constraints.
ManiSkill-HAB~\cite{shukla2025maniskill} provides a benchmark for this setting,
while HoMeR~\cite{sundaresan2025homer} and
MoManipVLA~\cite{wu2025momanipvla} study hybrid whole-body control and explicit
base--arm optimization.
Unlike methods that use trajectories as external sketches, visual subgoals, or
inputs to a separate optimizer, DT jointly generates a planned 3D
end-effector trajectory and a unified whole-body action chunk.

\subsection{World Action Models}
\label{sec:related:worldmodel}
World models predict action-conditioned futures for planning and policy
learning.
DINO-WM~\cite{zhou2024dino} predicts future pretrained visual features, and
the V-JEPA series~\cite{assran2025v,mur2026v} learns latent video
representations for action-conditioned planning.
Unified Video Action Model~\cite{li2025unified} and Unified World
Models~\cite{zhu2025unified} jointly model video and action dynamics.

Recent World Action Models further integrate future prediction with robot
control.
DreamZero~\cite{ye2026world} jointly predicts future visual states and actions,
MotuBrain~\cite{team2026motubrain} unifies video generation, inverse dynamics,
and action prediction, and GigaWorld-Policy~\cite{ye2026gigaworld} uses visual
future prediction as training supervision while supporting efficient
action-only inference.
In contrast to these image- or video-space models, DT predicts the compact
task-space trajectory induced by each candidate action and directly compares
it with the VLA-planned trajectory for test-time refinement.
\section{Method}
\label{sec:method}

DreamTrajectory, hereafter referred to as DT, is illustrated in
Figure~\ref{fig:dreamtrajectory}.
DT is motivated by a simple observation: in mobile manipulation, an action is
reliable only when it is consistent with the intended end-effector motion and
can be executed under the current scene geometry and contact conditions.
Based on this observation, we introduce end-effector trajectory guidance as an explicit
motion-level signal.

DT contains two main components.
First, a trajectory-guided VLA jointly predicts a future end-effector
trajectory and the whole-body action chunk intended to realize it
(Section~\ref{sec:method:tgag}).
Second, a lightweight action-conditioned trajectory world model is used at test
time (Section~\ref{sec:method:refine}) to refine the predicted action chunk.
The VLA therefore generates the intended motion, while the lightweight world
model evaluates whether a candidate action is likely to realize that motion
before execution.

\subsection{Problem Formulation}
\label{sec:method:problem}

We consider whole-body language-conditioned manipulation, where a mobile
manipulator must coordinate its base and arm to complete complex tasks.
At each replanning step $t$, the robot receives egocentric visual inputs $o_t$, a proprioceptive state $s_t$,
and a language instruction $\ell$ describing the task objective.
The goal is to learn a policy that predicts a future action chunk over a
horizon $H$, denoted by $a_{t:t+H-1}$.
Each action contains the low-level control commands required to coordinate the
mobile base and manipulator.

Formally, a standard VLA models the action distribution conditioned on the
current observation, proprioceptive state, and language instruction:
\begin{equation}
    \hat{a}_{t:t+H-1}
    \sim
    \pi_\theta
    \left(
        a_{t:t+H-1}
        \mid o_t, s_t, \ell
    \right).
\end{equation}

\subsection{Trajectory-Guided Action Generation}
\label{sec:method:tgag}

Following the flow-matching formulation of
$\pi_0$~\cite{black2024pi_0}, DT augments whole-body action generation with an
explicit task-space trajectory, providing structured motion guidance under the
large and kinodynamically constrained action space of mobile manipulation.

We represent the future end-effector motion as
$\tau_{t:t+H-1}=(\tau_t,\ldots,\tau_{t+H-1})$, where each waypoint is a
7D pose:
\begin{equation}
\tau_{t+h}
=
\left[
p_{t+h}^{B_t},
q_{t+h}^{B_t}
\right]
\in\mathbb{R}^{7}.
\end{equation}
Here, $p_{t+h}^{B_t}\in\mathbb{R}^{3}$ and
$q_{t+h}^{B_t}\in\mathbb{R}^{4}$ denote end-effector position and orientation in the
chunk-local frame $B_t$, anchored at the current base pose and fixed over the
prediction horizon.
This representation captures the combined effect of future base and arm
motion. The trajectory and action chunks share horizon $H$ and are jointly generated
by a dual-stream action expert:
\begin{equation}
    \left(
        \hat{\tau}_{t:t+H-1},
        \hat{a}_{t:t+H-1}
    \right)
    \sim
    \pi_\theta
    \left(
        \tau_{t:t+H-1},
        a_{t:t+H-1}
        \mid o_t, s_t, \ell
    \right).
\end{equation}

DT models the joint generation process using conditional flow matching.
Given a training trajectory chunk $\tau$ and action chunk $a$, we independently
sample Gaussian noise $\epsilon_\tau,\epsilon_a\sim\mathcal{N}(0,I)$ and use
a shared flow time $\sigma\sim\mathcal{U}(0,1)$ for the two streams.
The interpolated variables are
\begin{equation}
    \tau^\sigma
    =
    \sigma\epsilon_\tau+(1-\sigma)\tau,
    \qquad
    a^\sigma
    =
    \sigma\epsilon_a+(1-\sigma)a.
    \label{eq:joint_flow_interpolation}
\end{equation}
Using the same flow time allows the trajectory and action denoising processes
to evolve synchronously.

A shared Transformer predicts the trajectory and action velocity fields,
denoted by $v_\tau$ and $v_a$, respectively.
The trajectory-guided VLA is trained using the joint flow-matching objective
\begin{equation}
    \mathcal{L}_{\mathrm{VLA}}
    =
    \mathbb{E}
    \left[
        \lambda_\tau\|v_\tau-u_\tau\|_2^2
        +
        \lambda_a\|v_a-u_a\|_2^2
    \right].
    \label{eq:vla_flow_loss}
\end{equation}
where $u_\tau=\epsilon_\tau-\tau$ and
$u_a=\epsilon_a-a$ are the corresponding target velocities.

We impose an asymmetric dependency between the two output streams using
group-causal attention.
Trajectory tokens attend to the multimodal prefix and previous trajectory
tokens, whereas action tokens additionally attend to the complete trajectory
stream and previous action tokens.
This directional dependency allows the planned trajectory to guide action
denoising while preventing action information from leaking back into trajectory
generation.

At inference time, DT starts from independently sampled trajectory and action
noise and jointly integrates the learned velocity fields from $\sigma=1$ to
$\sigma=0$ using Euler steps.
The resulting samples form the planned trajectory
$\hat{\tau}_{t:t+H-1}$ and the corresponding whole-body action chunk
$\hat{a}_{t:t+H-1}$.
The target-velocity derivation, exact attention mask, and inference equations
are provided in supplementary materials.

\label{sec:experiments:main}
\begin{table*}[t]
\centering
\caption{Success rates (\%) on the MS-HAB simulator. Each method is evaluated for 100 episodes per task.}
\label{tab:sim_baselines}
\begin{tabular}{lccccccc}
\toprule
Method
& Pick Apple
& Pick Bowl
& Open Fridge
& Close Fridge
& Open Counter
& Close Counter
& Avg. \\
\midrule
ACT
& 1.0 & 0.0 & 28.0 & 24.0 & 22.0 & \textbf{93.0} & 28.0 \\
Diffusion Policy
& 19.0 & 22.0 & 19.0 & \textbf{61.0} & 17.0 & 28.0 & 27.7 \\
RDT-1B
& 0.0 & 0.0 & 5.0 & 14.0 & 0.0 & 33.0 & 8.7 \\
GR00T N1
& 3.0 & 1.0 & 43.0 & 16.0 & 5.0 & 59.0 & 21.2 \\
$\pi_{0.5}$
& 37.0 & 19.0 & 5.0 & 8.0 & 87.0 & 38.0 & 32.3 \\
DT (ours)
& \textbf{39.0} & \textbf{35.0} & \textbf{51.0}
& 33.0 & \textbf{91.0} & 80.0 & \textbf{54.8} \\
\bottomrule
\end{tabular}
\end{table*}

\begin{table*}[t]
\centering
\setlength{\tabcolsep}{4pt}
\caption{Ablation analysis on MS-HAB. Each method is evaluated for 100 episodes per task.}
\label{tab:mshab_ablation}
\begin{tabular}{@{}lccccccc@{}}
\toprule
Method
& \shortstack{Pick\\Apple}
& \shortstack{Pick\\Bowl}
& \shortstack{Open\\Fridge}
& \shortstack{Close\\Fridge}
& \shortstack{Open\\Counter}
& \shortstack{Close\\Counter}
& Avg. \\
\midrule
DT w/o trajectory guidance
& 37.0 & 19.0 & 5.0 & 8.0 & 87.0 & 38.0 & 32.3 \\
DT w/o refiner
& 34.0 & 33.0 & 44.0 & 22.0 & 80.0 & 72.0 & 47.5 \\
DT
& \textbf{39.0} & \textbf{35.0} & \textbf{51.0}
& \textbf{33.0} & \textbf{91.0} & \textbf{80.0}
& \textbf{54.8} \\
\bottomrule
\end{tabular}
\end{table*}

% \begin{figure*}[]
%     \centering
%     \begin{subfigure}[t]{0.31\linewidth}
%         \centering
%         \includegraphics[width=\linewidth]{Figures/pick_fruit_frame40.png}
%         \caption{Fruit pick-and-place.}
%         \label{fig:real_pick_fruit}
%     \end{subfigure}
%     \hfill
%     \begin{subfigure}[t]{0.31\linewidth}
%         \centering
%         \includegraphics[width=\linewidth]{Figures/open_drawer_frame40.png}
%         \caption{Opening a drawer.}
%         \label{fig:real_open_drawer}
%     \end{subfigure}
%     \hfill
%     \begin{subfigure}[t]{0.31\linewidth}
%         \centering
%         \includegraphics[width=\linewidth]{Figures/close_drawer_frame40.png}
%         \caption{Closing a drawer.}
%         \label{fig:real_close_drawer}
%     \end{subfigure}
%     \caption{Real-world mobile manipulation tasks.}
%     \label{fig:real_world_tasks}
% \end{figure*}

% \begin{figure*}[!t]
%     \centering
%     \includegraphics[width=0.75\textwidth]{Figures/imgdata.pdf}
%     \caption{
% Simulation and real-world mobile manipulation tasks. All tasks require the coordination of base and arm motion.
% }
%     \label{fig:tasks}
% \end{figure*}

\begin{figure*}[!t]
    \centering

    % ==================== Simulation ====================
    \makebox[\textwidth]{%
        \rule{0.38\textwidth}{0.8pt}
        \hspace{0.8em}
        {\large\bfseries Simulation}
        \hspace{0.8em}
        \rule{0.38\textwidth}{0.8pt}
    }

    \medskip

    % Simulation: Pick the Fruit
    \begin{minipage}[t]{0.32\textwidth}
        \centering
        {\large Pick the Fruit\par}
        \medskip
        \includegraphics[
            width=\linewidth,
            height=0.115\textheight,
            keepaspectratio
        ]{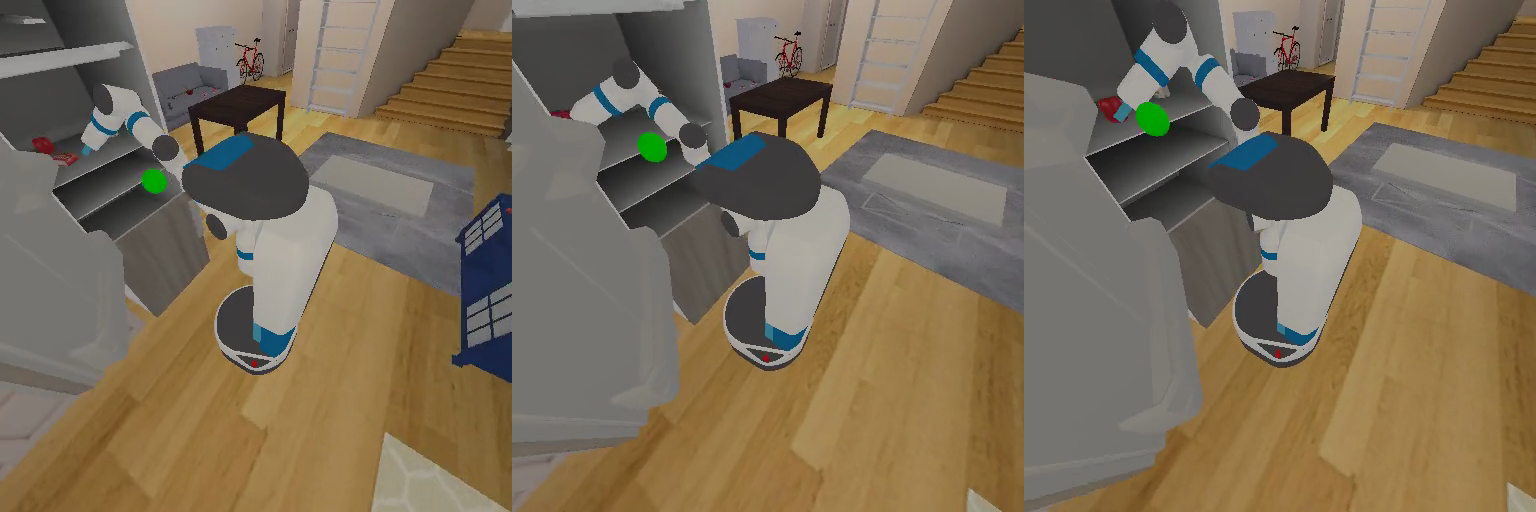}
    \end{minipage}
    \hfill
    % Simulation: Close the Fridge
    \begin{minipage}[t]{0.32\textwidth}
        \centering
        {\large Close the Fridge\par}
        \medskip
        \includegraphics[
            width=\linewidth,
            height=0.115\textheight,
            keepaspectratio
        ]{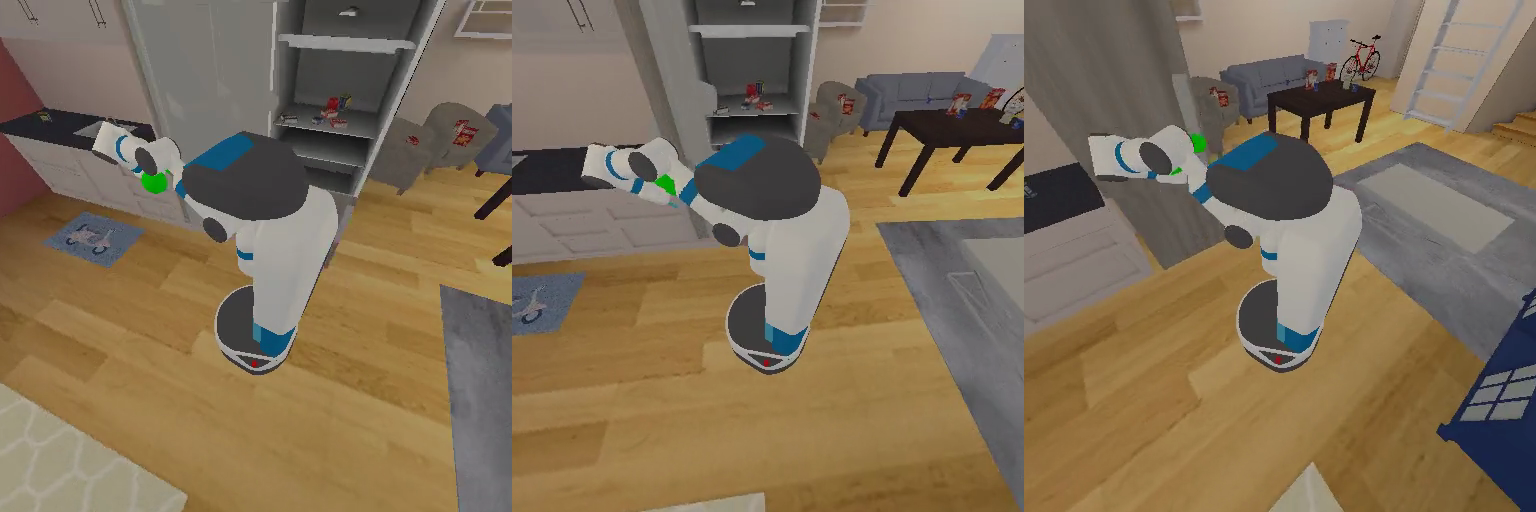}
    \end{minipage}
    \hfill
    % Simulation: Open the Drawer
    \begin{minipage}[t]{0.32\textwidth}
        \centering
        {\large Open the Drawer\par}
        \medskip
        \includegraphics[
            width=\linewidth,
            height=0.115\textheight,
            keepaspectratio
        ]{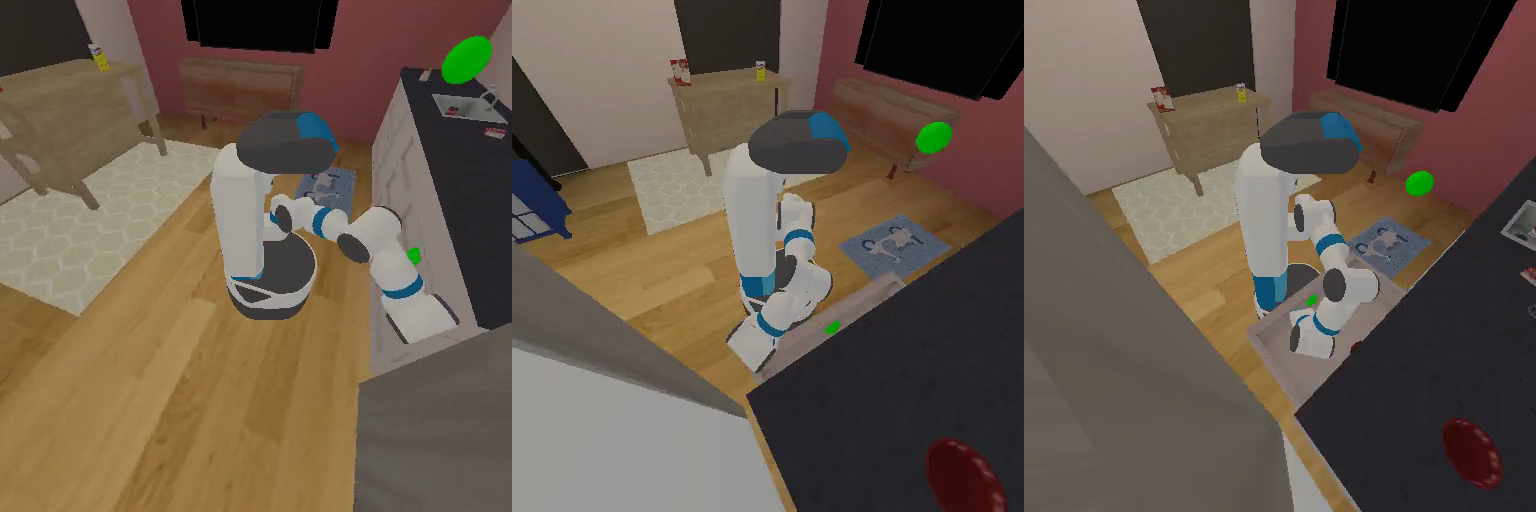}
    \end{minipage}

    \medskip

    % ==================== Real World ====================
    \makebox[\textwidth]{%
        \rule{0.37\textwidth}{0.8pt}
        \hspace{0.8em}
        {\large\bfseries Real World}
        \hspace{0.8em}
        \rule{0.37\textwidth}{0.8pt}
    }

    \medskip

    % Real World: Pick the Fruit
    \begin{minipage}[t]{0.32\textwidth}
        \centering
        {\large Pick the Fruit\par}
        \medskip
        \includegraphics[
            width=\linewidth,
            height=0.105\textheight,
            keepaspectratio
        ]{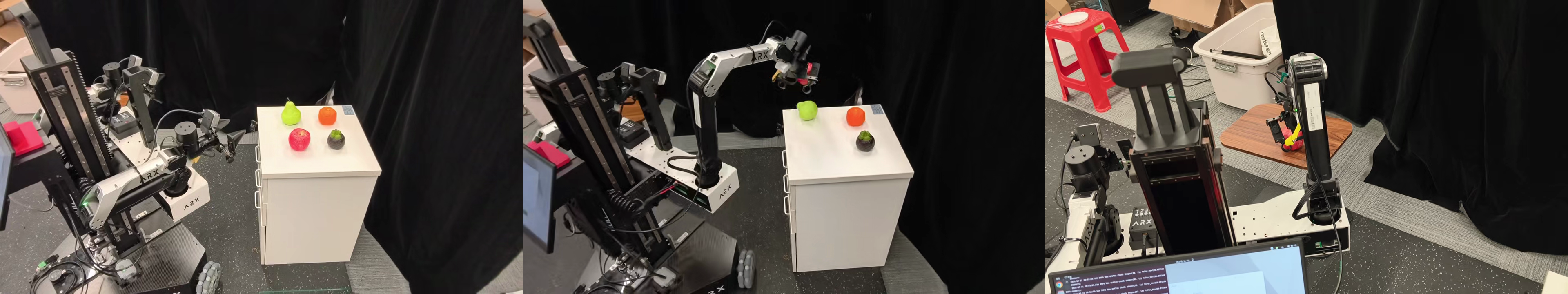}
    \end{minipage}
    \hfill
    % Real World: Close the Drawer
    \begin{minipage}[t]{0.32\textwidth}
        \centering
        {\large Close the Drawer\par}
        \medskip
        \includegraphics[
            width=\linewidth,
            height=0.105\textheight,
            keepaspectratio
        ]{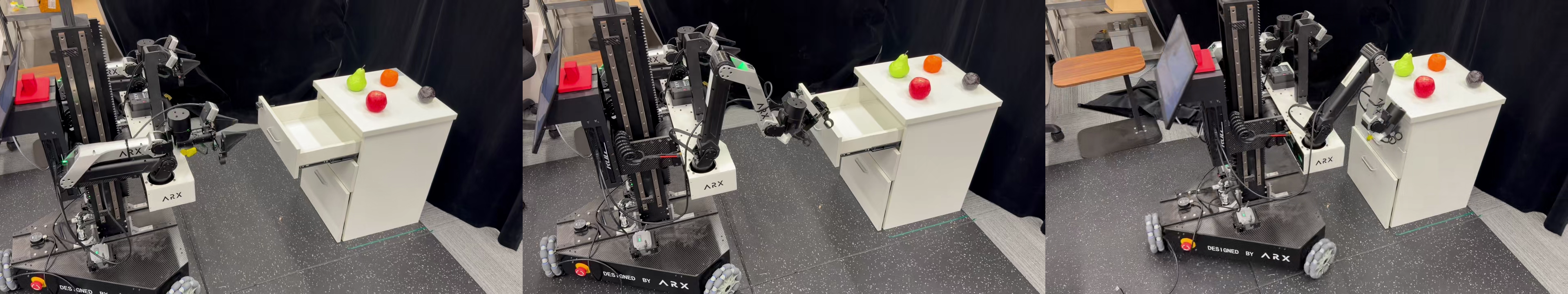}
    \end{minipage}
    \hfill
    % Real World: Open the Drawer
    \begin{minipage}[t]{0.32\textwidth}
        \centering
        {\large Open the Drawer\par}
        \medskip
        \includegraphics[
            width=\linewidth,
            height=0.105\textheight,
            keepaspectratio
        ]{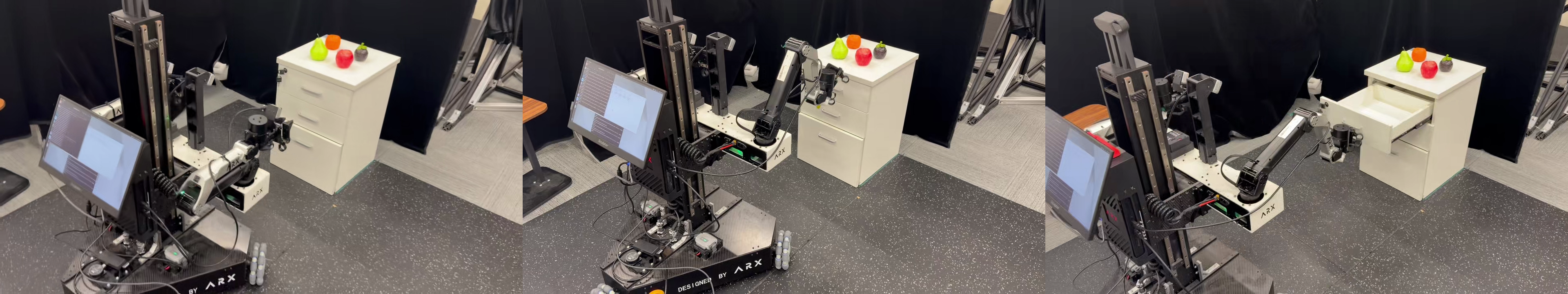}
    \end{minipage}

    \caption{
        Simulated and real-world mobile manipulation tasks.
        All tasks require coordinated base and arm motion.
    }
    \label{fig:tasks}
\end{figure*}

\begin{table}[t]
\centering
\small
\caption{Real-world task success rates (\%). Each method is evaluated for
20 episodes per task.}
\label{tab:real_world}
\begin{tabular}{@{}lcccc@{}}
\toprule
Method
& \shortstack{Fruit\\Pick-and-Place}
& \shortstack{Open\\Drawer}
& \shortstack{Close\\Drawer}
& Avg. \\
\midrule
$\pi_{0.5}$
& 45.0 & 60.0 & 85.0 & 63.3 \\
DT w/o refiner
& 70.0 & 75.0 & 100.0 & 81.7 \\
DT
& \textbf{80.0} & \textbf{90.0} & \textbf{100.0} & \textbf{90.0} \\
\bottomrule
\end{tabular}
\end{table}

\subsection{Action Refinement with a Trajectory World Model}
\label{sec:method:refine}

To reduce plan-to-execution deviations caused by contacts, collisions, and
control errors, DT uses a trajectory world model to refine the VLA-predicted
action chunk before execution.

\begin{table*}[t]
\centering
\small
\setlength{\tabcolsep}{5pt}
\caption{
World Model Architecture Comparison
}
\label{tab:wm_architecture}
\begin{tabular}{lcccc}
\toprule
Model
& xyz ADE (m) $\downarrow$
& xyz FDE (m) $\downarrow$
& Angular ADE ($^\circ$) $\downarrow$
& Geodesic ADE $\downarrow$
\\
\midrule
Analytical FK (Open-Loop)
& 0.241
& 0.345
& 27.0
& 0.057\\
\midrule
Diffusion
& 0.061
& 0.086
& 8.1
& \textbf{0.014}
\\
One-shot Transformer
& 0.036
& 0.049
& 7.7
& 0.016
\\
Cross Attention
& 0.033
& 0.045
& 7.4
& 0.016
\\
GRU (Used in DT)
& \textbf{0.028}
& \textbf{0.035}
& \textbf{6.2}
& \textbf{0.014}
\\
\bottomrule
\end{tabular}
\end{table*}

\paragraph{Action-conditioned trajectory world model.}
Given dual-view observation
$o_t=(I_t^{\mathrm{head}},I_t^{\mathrm{wrist}})$,
state $s_t$, and candidate action chunk
$a_{t:t+H-1}\in\mathbb{R}^{H\times d_a}$,
the recurrent world model predicts its induced trajectory:
\begin{equation}
    \tilde{\tau}(a)
    =
    \mathcal{W}_{\phi}
    \left(
        o_t,
        s_t,
        a_{t:t+H-1}
    \right).
\end{equation}
Here, $\tilde{\tau}(a)\in\mathbb{R}^{H\times7}$ is the predicted task-space
execution outcome rather than the VLA-planned trajectory.
Trained on recorded interaction outcomes, the model captures contact- and
control-induced deviations beyond nominal kinematics.

The predicted trajectory is compared with the VLA plan
$\hat{\tau}$ during candidate selection.
At each replanning step, DT constructs the candidate set in denormalized action
space. It retains the original chunk $\hat{a}$ and independently samples
$N-1$ perturbation chunks $\epsilon_i$, where each action dimension follows an
independent AR(1) Gaussian process with marginal standard deviation
$\sigma=0.05$ and lag-one correlation $\rho=0.9$.
With $N=30$, the candidate set is
\begin{equation}
    \mathcal{C}
    =
    \left\{\hat{a}\right\}
    \cup
    \left\{
        \hat{a}+\epsilon_i
        \,\middle|\,
        i=1,\ldots,N-1
    \right\}.
\end{equation}

Next, the world model evaluates the induced trajectory of every candidate in
$\mathcal{C}$ in parallel.
Language is omitted because the world model predicts task-agnostic task-space execution outcomes; task intent is represented by the planned trajectory.

Finally, DT selects the candidate that best balances trajectory agreement and
motion smoothness:
\begin{equation}
    \tilde{a}
    =
    \arg\max_{a\in\mathcal{C}}
    \left[
        \lambda S_{\mathrm{traj}}(a)
        +
        (1-\lambda)\eta S_{\mathrm{smooth}}(a)
    \right].
    \label{eq:refinement_objective}
\end{equation}
Here, $S_{\mathrm{traj}}$ measures planned--induced trajectory agreement and
$S_{\mathrm{smooth}}$ penalizes abrupt motion.
We use $\lambda=0.5$ and $\eta=10^{-3}$; detailed definitions are provided in
supplementary materials.

% \begin{algorithm}[t]
% \caption{Search--Predict--Score Action Refinement}
% \label{alg:action_refinement}
% \begin{algorithmic}[1]
% \REQUIRE $o_t,s_t,\ell,\pi_\theta,\mathcal{W},N,
% \mathcal{P}_\epsilon,\lambda,\eta$
% \ENSURE Refined action chunk $\tilde{a}$

% \STATE $(\hat{\tau},\hat{a})
% \sim\pi_\theta(\tau,a\mid o_t,s_t,\ell)$
% \STATE $\mathcal{C}\leftarrow\emptyset$

% \FOR{$i=0$ \TO $N-1$}
%     \IF{$i=0$}
%         \STATE $\epsilon_i\leftarrow 0$
%     \ELSE
%         \STATE $\epsilon_i\sim\mathcal{P}_\epsilon$
%     \ENDIF
%     \STATE $a_i\leftarrow
%     \operatorname{clip}(\hat{a}+\epsilon_i)$
%     \STATE $\mathcal{C}\leftarrow\mathcal{C}\cup\{a_i\}$
% \ENDFOR

% \FORALL{$a\in\mathcal{C}$}
%     \STATE $\tilde{\tau}(a)\leftarrow
%     \mathcal{W}(o_t,s_t,a)$
%     \STATE $S(a)\leftarrow
%     \lambda S_{\mathrm{traj}}(a)
%     +(1-\lambda)\eta S_{\mathrm{smooth}}(a)$
% \ENDFOR

% \STATE $\tilde{a}\leftarrow
% \arg\max_{a\in\mathcal{C}}S(a)$
% \RETURN $\tilde{a}$
% \end{algorithmic}
% \end{algorithm}

\subsection{Training}
DT is trained in two separately optimized stages.

\textbf{Stage I: VLA Training.}
We initialize the VLA backbone using the pretrained weights of $\pi_{0.5}$ and
fine-tune it on expert demonstrations.
Each training sample contains
$(o_t,s_t,\ell,\tau_{t:t+H-1},a_{t:t+H-1})$.
The trajectory labels contain future 7D end-effector poses.
The end-effector poses are expressed in the robot base frame at the current
replanning step $t$, which remains fixed throughout the prediction horizon.
As a result, the trajectory captures the combined effect of future base and
arm motion.

\textbf{Stage II: World Model Training.}
We train the trajectory world model using an additional dataset of
robot--environment interactions containing both successful and failed trials.
For each trial, a training sample is constructed as
$d_s=(I_t^{\mathrm{head}},I_t^{\mathrm{wrist}},s_t,
a_{t:t+H-1},\tau^{\mathrm{exec}}_{t:t+H-1})$.
Specifically, we integrate the logged planar base-velocity channels over the
action chunk and transform the recorded end-effector poses from the
instantaneous robot-base frame into a chunk-local coordinate frame anchored at
the base pose at time $t$.
A one-step temporal offset is used such that the target waypoint at step $h$
corresponds to the end-effector pose observed after executing action
$a_{t+h}$.
Thus, $\tau^{\mathrm{exec}}$ represents the physical trajectory induced by the
recorded action chunk, rather than an open-loop planned trajectory.

We construct dense training samples using sliding windows over the recorded
episodes.
The action and trajectory dimensions are independently standardized using
training-set statistics before world-model training.
The trajectory world model is optimized with a supervised trajectory
prediction objective, whose detailed formulation is provided in
supplementary materials.

At deployment, the VLA and trajectory world model are combined only through
the test-time refinement procedure described in
Section~\ref{sec:method:refine}.
Additional architectural details and hyperparameters are provided in the
supplementary materials.

\section{Experiments}
\label{sec:experiments}

We evaluate DreamTrajectory on simulated and real-world mobile manipulation tasks.
The experiments address three questions:
(i) how DT compares with existing baselines on MS-HAB,
(ii) whether explicit trajectory guidance improves whole-body action generation,
(iii) whether the same benefits hold on a physical mobile manipulator.

% The experiments are organized around four questions: (i)~how DreamTrajectory compares with existing baselines on MS-HAB, (ii)~whether an explicit end-effector trajectory improves whole-body action generation, (iii)~whether the trajectory world model further closes the plan-to-execution gap at inference time, and (iv)~whether the same benefits transfer to a physical mobile manipulator.

\subsection{Experimental Setup}
\label{sec:experiments:setup}

\textbf{Simulation benchmark.}
We define mobile manipulation as requiring concurrent base and arm motion,
rather than sequential navigation followed by fixed-base manipulation, since
such simultaneous control directly evaluates whole-body coordination.
Among the publicly available benchmarks we surveyed, ManiSkill-HAB
(MS-HAB)~\cite{shukla2025maniskill} best satisfies this criterion by providing
contact-rich tasks with frequent simultaneous base--arm motion.
We therefore evaluate DT on its \texttt{set\_table} suite using a Fetch mobile
manipulator.
The evaluation includes six subtasks:
\texttt{pick apple}, \texttt{pick bowl}, \texttt{open fridge},
\texttt{close fridge}, \texttt{open counter}, and
\texttt{close counter}.
These tasks require concurrent base--arm control under changing viewpoints and
contact conditions.
The reinforcement-learning demonstrations contain diverse whole-body
trajectories with frequent simultaneous base and arm motion.

\textbf{Real-world setup.}
We conduct real-world experiments on an ARX LIFT mobile manipulator using fruit
pick-and-place, drawer opening, and drawer closing.
The fruit task includes four objects: apple, orange, pear, and mangosteen.
All tasks require base adjustment during manipulation because of the limited
arm workspace.

\textbf{Baselines and variants.}
Table~\ref{tab:sim_baselines} compares DT with representative learning-based
robot policies.
Table~\ref{tab:mshab_ablation} evaluates three DT variants:
\textbf{DT w/o trajectory guidance} directly predicts whole-body actions;
\textbf{DT w/o refiner} retains trajectory-guided generation but executes the
initial action chunk directly; and
\textbf{DT} additionally applies trajectory world-model refinement over 30
candidates at each replanning step.
All variants share the same demonstrations, inputs, action representation, and
evaluation protocol.
Future-prediction VLAs are excluded from quantitative comparison because their
fixed-base embodiments and interfaces are incompatible with MS-HAB; their
designs are compared qualitatively in
Table~\ref{tab:qualitative_comparison}.

\textbf{Trajectory world model.}
The GRU-based world model takes the dual-view observation, proprioceptive
state, and an $H=16$ candidate action chunk as input and predicts a sequence
of 7D end-effector poses in the chunk-local frame.
DT evaluates all 30 candidates in parallel over the batch dimension.
Architecture and training details are provided in
supplementary materials.

\textbf{Metric.}
We report task success rate over 100 episodes per simulation subtask, yielding
600 episodes per method.
A rollout succeeds if the task goal is completed within the episode horizon,
and the overall score is averaged across the six subtasks.
For the real-world evaluation, each method is tested over 20 episodes per task,
and the average is computed across the three tasks.

\subsection{Simulation Experiment Results}
Table~\ref{tab:sim_baselines} summarizes the simulation comparison results.
DreamTrajectory achieves the highest average success rate among the evaluated end-to-end baselines, with particularly large gains on several contact-rich fridge and counter interaction tasks.
These results support our hypothesis that combining trajectory-guided action generation with test-time search--predict--score refinement is more effective than directly predicting whole-body actions.

\subsection{Ablation Study}
\label{sec:experiments:analysis}

Table~\ref{tab:mshab_ablation} isolates the contributions of trajectory
guidance and world-model refinement.
Adding trajectory guidance to the action-only $\pi_{0.5}$ baseline improves
average success from $32.3\%$ to $47.5\%$.
The largest gains occur on contact-rich articulated-object tasks:
\texttt{open fridge} increases from $5.0\%$ to $44.0\%$, and
\texttt{close counter} from $38.0\%$ to $72.0\%$.
The slight decreases on \texttt{pick apple} and \texttt{open counter} indicate
that trajectory guidance primarily benefits tasks requiring precise
whole-body contact coordination.

World-model refinement further raises average success to $54.8\%$, improving
all six subtasks and adding $7.3$ percentage points over the trajectory-guided
policy.
For example, \texttt{close counter} improves from $72.0\%$ to $80.0\%$,
showing that refinement also corrects residual errors when the initial policy
is already strong.
Overall, the $22.5$-point gain over the action-only baseline consists of
$15.2$ points from trajectory guidance and $7.3$ points from refinement,
demonstrating that the two components are complementary.

\subsection{World-Model Architecture Ablation}
\label{sec:experiments:wm_architecture}

We compare an analytical open-loop forward kinematics model (denoted as Analytical FK in Table~\ref{tab:wm_architecture}) rollout with four learned trajectory
predictors: One-shot Transformer, Cross Attention, GRU, and Diffusion.
All learned models take the same dual-view observation, proprioceptive state,
and action chunk as input and predict an $H=16$ end-effector trajectory in a
chunk-local world frame.
They use identical trajectory targets and are evaluated on 4,096 trajectories.
Predicted quaternions are unit-normalized and sign-aligned with the ground
truth before evaluation.

As shown in Table~\ref{tab:wm_architecture}, all learned models substantially
outperform FK, which cannot capture contacts, controller errors, or other
execution deviations beyond nominal kinematics.
GRU achieves the best overall accuracy, with an xyz ADE of $0.028$\,m, an xyz
FDE of $0.035$\,m, and an angular ADE of $6.2^\circ$.
Diffusion matches GRU in geodesic ADE but produces larger position and angular
errors, while Cross Attention and One-shot Transformer remain competitive but
less accurate.
These results motivate our selection of the GRU predictor, which achieves the best overall trajectory-prediction accuracy for the $H=16$ horizon.

\subsection{Computational Efficiency}
\label{sec:experiments:efficiency}

We analyze the additional computational cost introduced by trajectory-guided
action generation and test-time refinement. Inference latency is measured on a
single RTX 4090 using BF16 precision.

\begin{table}[t]
\centering
\small
\caption{Additional computational cost relative to the $\pi_{0.5}$ VLA.}
\label{tab:computational_efficiency}
\begin{tabular}{@{}lcc@{}}
\toprule
Component & Added params & Added latency \\
\midrule
Trajectory head
& 0.017M ($<0.01\%$)
& 3.65\,ms \\
Trajectory world model
& 49.02M (1.40\%)
& 8.10\,ms \\
\midrule
Complete DT
& 49.04M (1.40\%)
& 11.75\,ms \\
\bottomrule
\end{tabular}
\end{table}

As shown in Table~\ref{tab:computational_efficiency}, the trajectory head adds
only 0.017M parameters, making its parameter overhead negligible relative to
the underlying $\pi_{0.5}$ VLA. The trajectory world model adds 49.02M
parameters, corresponding to only 1.40\% of the 3.5B-parameter VLA.
Overall, the complete DT framework introduces only 49.04M additional
parameters (1.40\%) and 11.75\,ms of latency per replanning step. With this limited computational overhead, the complete DT framework provides a further improvement in average success rate
from $47.5\%$ without refinement to $54.8\%$ with the complete DT framework.

\subsection{Real-World Experiments}
\label{sec:experiments:real}
We evaluate DT on three real-world tasks using an ARX LIFT mobile
manipulator: fruit pick-and-place, drawer opening, and drawer closing
(Figure~\ref{fig:tasks}). Each method is evaluated for 20 episodes per task
under the same protocol; additional setup and success-criterion details are
provided in the supplementary materials.

As shown in Table~\ref{tab:real_world}, trajectory guidance improves average
success from $63.3\%$ to $81.7\%$, and refinement further improves it to
$90.0\%$. The complete DT particularly improves fruit pick-and-place from
$45.0\%$ to $80.0\%$ and drawer opening from $60.0\%$ to $90.0\%$.
Trajectory guidance already achieves $100.0\%$ on drawer closing. The gains therefore suggest that the planned end-effector trajectory provides an effective task-space guide beyond direct action prediction, while refinement corrects residual plan--execution deviations before execution.

\section{Conclusion}
\label{sec:conclusion}

In this paper, we present DreamTrajectory, a mobile manipulation framework that uses end-effector trajectories as an explicit interface between perception and whole-body control. The policy jointly denoises a planned trajectory and a whole-body action chunk, providing task-space guidance for high-dimensional base--arm action generation. At inference, a lightweight trajectory world model predicts the motion induced by each candidate action chunk, while a search--predict--score procedure selects the candidate whose predicted trajectory best matches the plan. By sharing the same task-space representation across action generation and consequence prediction, DreamTrajectory enables candidate actions to be evaluated against the policy's intended motion before execution. On the MS-HAB \texttt{set\_table} suite, trajectory guidance increases mean success from $32.3\%$ to $47.5\%$ over an action-only $\pi_{0.5}$ baseline, and test-time refinement further improves it to $54.8\%$, with the largest gains on contact-rich articulated-object tasks. On three real-world tasks, the corresponding success rates increase from $63.3\%$ to $81.7\%$ and then to $90.0\%$. These results show that trajectory-guided generation and planned--induced trajectory alignment provide an effective and computationally lightweight approach to coordinated mobile manipulation.

\bibliography{aaai2027}

\end{document}